\documentclass[conference]{IEEEtran}
\usepackage{amsmath,amssymb,graphicx,booktabs,cite}
\usepackage{amsthm}
\theoremstyle{definition}
\newtheorem{definition}{Definition}
\newtheorem{proposition}{Proposition}
\graphicspath{{figs/}}
\begin{document}
\title{LOOP-MAC: Learning-Based Multi-Agent Coordination with Guaranteed Local Feasibility in a Virtual Power Plant\thanks{This work has been submitted to the IEEE for possible publication. Copyright may be transferred without notice, after which this version may no longer be accessible.}}

\author{\IEEEauthorblockN{Meiyi Li}
\IEEEauthorblockA{Louisiana State University\\Baton Rouge, LA, USA\\ mli30@lsu.edu}
\and
\IEEEauthorblockN{Lucas Ruda}
\IEEEauthorblockA{The University of Texas at Austin\\Austin, TX, USA\\ lucasruda@utexas.edu}
\and
\IEEEauthorblockN{Javad Mohammadi}
\IEEEauthorblockA{The University of Texas at Austin\\Austin, TX, USA\\ javadm@utexas.edu}
}
\IEEEoverridecommandlockouts
\IEEEaftertitletext{\vspace{-2.5\baselineskip}}
\maketitle

\begin{abstract}

By aggregating distributed energy resources (DERs) into a single dispatchable resource, virtual power plants (VPPs) enable the grid to use the flexibility of numerous small-scale DERs. Delivering an aggregate VPP commitment requires multi-agent coordination, where agents manage local DER models and operating constraints while jointly meeting a power commitment within limited communication rounds. Classical distributed optimization is computationally intensive, requiring iterative local optimization and repeated communication to reach consensus. Learning-based methods can accelerate coordination but often lack guarantees of local feasibility and accurate commitment tracking.
This paper proposes LOOP-MAC (Learning to Optimize the Optimization Process for Multi-Agent Coordination), a fully decentralized learning-based framework that ensures local feasibility by construction while efficiently reducing the aggregate commitment gap. LOOP-MAC employs a two-stage decision process. Learned coordination uses neural optimizers equipped with a gauge layer to produce locally feasible schedules through a few rounds of message exchange. Local adjustment then rebalances each schedule toward feasible flexibility extremes using a commitment-gap estimate inferred from messages already received. The adjusted schedule remains locally feasible as a convex combination of feasible schedules, and the adjustment requires no additional communication round.  On a 100-agent VPP, five communication rounds followed by local adjustment reduce the mean commitment gap from 5.1\% to 0.75\% of the agents' aggregate absolute power and the optimality gap from 3.93\% to 3.07\% relative to learned coordination alone. Per-agent online computation takes approximately 1.1~ms, including the adjustment.

\end{abstract}


\section{Introduction}

\begin{figure*}[t]\centering
\includegraphics[width=\textwidth]{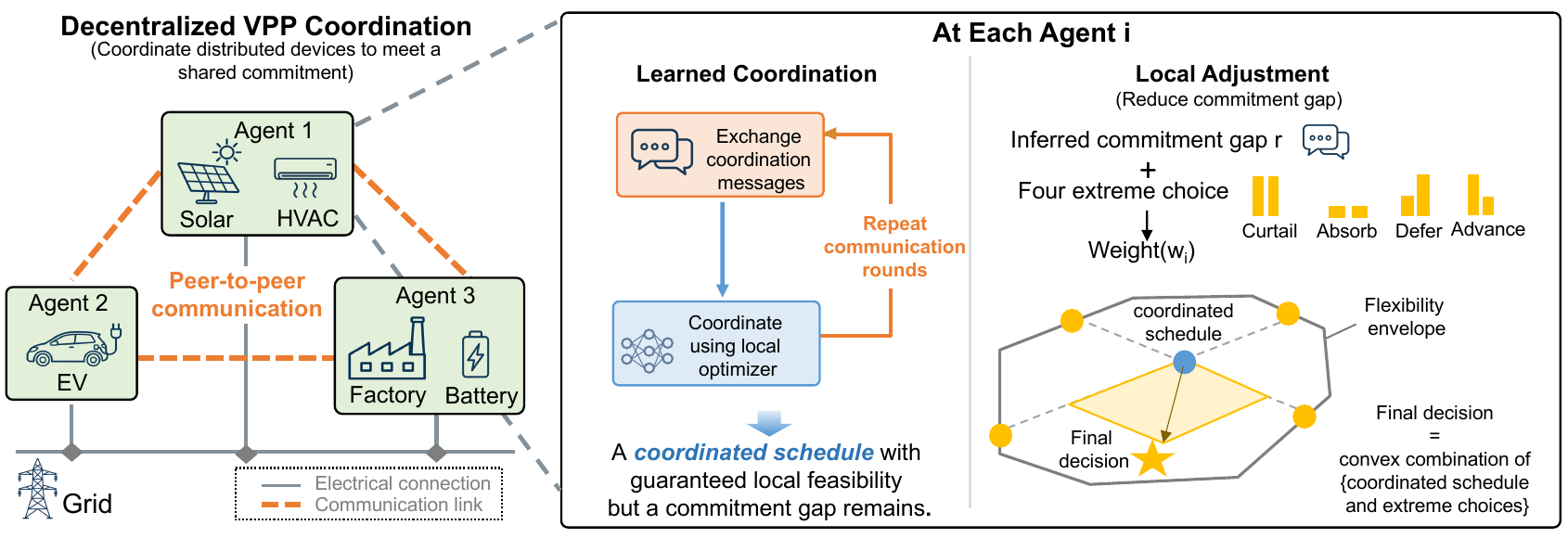}
\caption{LOOP-MAC: a fully decentralized learning-based framework for VPP commitment
delivery. Agents coordinate through a few rounds of message exchange and then decide
locally, so every schedule is locally feasible by construction and the aggregate
commitment gap stays small.}
\label{fig:method}
\end{figure*}

By aggregating distributed energy resources (DERs) into a single dispatchable resource, virtual power plants (VPPs) enable the grid to harness the flexibility of numerous small-scale DERs \cite{chen2023tracking}. To deliver an aggregate power commitment, a VPP must coordinate power schedules across heterogeneous DERs with diverse dynamics, operating constraints, and user preferences \cite{chen2024disagg,wang2026l2o}. Centralized optimization can determine these schedules but requires access to individual DER models and may face scalability challenges as the number of DERs grows. This motivates multi-agent coordination, in which agents retain their DER models and operating constraints locally while coordinating their power schedules to meet a shared commitment within a limited number of communication rounds.

Classical distributed optimization enables such coordination through message exchange without requiring agents to disclose their local models \cite{molzahn2017survey}. However, these methods can incur substantial computation and many communication rounds. For example, the alternating direction method of multipliers (ADMM) \cite{boyd2011admm} typically requires each agent to solve a constrained local optimization subproblem at every iteration, followed by repeated message exchanges to approach consensus \cite{zeng2022rl,feng2023admm}. Learning-accelerated variants reduce computational effort by warm-starting local solvers or coordination variables \cite{biagioni2022admm,mak2023admm}, while unfolded distributed optimizers aim to reduce the number of coordination iterations \cite{noah2025unfold,saravanos2025deep}. Nevertheless, these approaches retain elements of the iterative optimization structure.

Learning-based coordination offers an alternative by replacing iterative local optimization with neural optimizers and coordinating agents through only a few rounds of message exchange \cite{li2023l2o}. However, reliable coordination must address two distinct requirements: satisfying each agent's local operating constraints and tracking the aggregate power commitment that couples all agents. For local feasibility, incorporating constraint violations into the training loss \cite{liu2024loop} is straightforward but does not guarantee feasible decisions. Correction- and projection-based methods \cite{donti2021dc3,liang2023homeo} can enforce feasibility, but may require additional computational steps that diminish the speed advantage of neural optimization. In contrast, gauge maps \cite{li2023access,li2024gm} transform neural network outputs into locally feasible decisions through a closed-form, differentiable mapping, enabling feasibility by construction without iterative correction. Yet local feasibility alone does not ensure that the agents' aggregate power matches the VPP commitment. Existing learned coordinators typically address this coupling through training losses and limited message exchange, which can leave a remaining commitment gap. Primal-recovery schemes for dual methods reduce such coupling violations by convexly combining
several responses of each agent \cite{parshakova2025mra}, but the combination weights are
optimized jointly across agents, i.e., by a coordinator. Reducing this gap through additional coordination rounds would delay the schedule decision, motivating a mechanism that improves commitment tracking using information already available to each agent.

To address these challenges, this paper proposes LOOP-MAC, a fully decentralized learning-based framework that guarantees local feasibility while reducing the aggregate commitment gap without additional coordination rounds. LOOP-MAC employs a two-stage decision process. In the \emph{learned coordination} stage, agents exchange coordination messages and update their schedules using a neural optimizer equipped with a gauge map, ensuring that local operating constraints are satisfied by construction. The subsequent \emph{local adjustment} stage reduces the remaining commitment gap with each agent's own flexibility: the agent shifts its coordinated schedule toward the flexibility extremes (curtailment, absorption, deferral, or advance) that offset the gap, inferred from coordination messages already received, and the result stays locally feasible because it is a convex combination of feasible schedules.

\section{Problem Setting}\label{sec:setting}
The VPP coordinates the consumption and production of its DERs so that their aggregate power
meets a market commitment while every DER respects its physical and operational limits.

\textbf{Agents.} The VPP contains $N$ agents, indexed by $i$. Each agent represents a cluster of
heterogeneous DERs: flexible loads (FL), energy storage (ESS), HVAC, EV charging and PV.

\textbf{Centralized formulation.} Over $T$ dispatch windows, the coordination task is
\begin{subequations}\label{eq:central}
\begin{align}
\min_{\{u_i\}}\ & \textstyle\sum_{i=1}^{N} f_i(u_i) \label{eq:central_obj}\\
\text{s.t.}\ & u_i\in\mathcal{U}_i(x_i),\quad i=1,\dots,N, \label{eq:central_local}\\
& \textstyle\sum_{i=1}^{N} p_i = s . \label{eq:central_couple}
\end{align}
\end{subequations}
Here $u_i$ is agent $i$'s schedule, i.e., the power of each of its devices in each window, and
$x_i$ collects its local inputs (forecasts and initial states). The agent's net power
$p_i\in\mathbb{R}^T$ in window $t$ is
\begin{equation}
p_{i,t}=p^{\mathrm{PV}}_{i,t}+p^{\mathrm{dis}}_{i,t}-p^{\mathrm{ch}}_{i,t}
-p^{\mathrm{IL}}_{i,t}-p^{\mathrm{FL}}_{i,t}-p^{\mathrm{HVAC}}_{i,t}-p^{\mathrm{EV}}_{i,t},
\label{eq:netpower}
\end{equation}
where the terms are the power of PV, battery discharging and charging, inflexible load (a
forecast), flexible load, HVAC and EV charging; it is written compactly as $p_i=C_iu_i$ with a constant matrix $C_i$.
The local cost $f_i$ in \eqref{eq:central_obj} penalizes battery cycling, deviation of the
flexible load from its reference, and deviation of the indoor temperature from the preferred
setpoint. The local feasible set $\mathcal{U}_i(x_i)$ in \eqref{eq:central_local} collects the
device limits: PV output given by the irradiance forecast; ESS charging and discharging limits
and state-of-charge dynamics and bounds; the flexible-load operating range; the HVAC thermal
dynamics, comfort band and power rating; the EV charging range and required energy; and the
agent's net-power limit. Constraint \eqref{eq:central_couple} requires the aggregate net power
to meet the VPP commitment $s\in\mathbb{R}^T$ in every window. We use continuous device models with affine dynamics and linear limits, so $\mathcal{U}_i(x_i)$ is a convex polytope.

\textbf{Decentralized coordination and its challenges.} Each agent knows only its own $x_i$,
$f_i$ and $\mathcal{U}_i$, together with the commitment $s$, so \eqref{eq:central} is solved by the agents themselves: they
exchange coordination messages over communication rounds, and each decides its own schedule.
This raises two requirements. First, the schedule each agent outputs must satisfy its own
device limits \eqref{eq:central_local}, and computing it must be cheap enough to repeat every
round. Second, the agents' schedules, each decided locally, must jointly meet the commitment
\eqref{eq:central_couple}. Section~\ref{sec:method} addresses the first with learned
coordination and the second with a local adjustment step.

\section{Proposed Fully Decentralized Learning-Based Coordination}\label{sec:method}
LOOP-MAC has two stages: in learned coordination, each agent updates its schedule
with a neural optimizer whose output always stays within its limits; in local adjustment, each
agent estimates the remaining commitment gap from the messages already received and shifts its
schedule within its own flexibility to reduce it, with no extra communication round.
\subsection{Stage 1: Learned coordination}
Following \cite{li2023l2o}, agent $i$ receives coordination messages $m_i^k$ in round $k$ summarizing the other agents' previous-round power schedules and their estimates of shared scheduling decisions. These messages provide the context for aligning local schedules without exchanging device models. Every pair communicates because \eqref{eq:central_couple} couples all agents. Agent $i$ updates its schedule with a neural optimizer $\xi_i$,

\begin{equation}
u_i^k=\xi_i\big(x_i,m_i^k\big).
\label{eq:update}
\end{equation}
The neural optimizer has two parts: a neural network $\phi_i$ that produces a preliminary
decision, which may violate the local constraints, and a gauge layer that maps this
decision into $\mathcal{U}_i(x_i)$. The network outputs only the independent variables $z_i$
of the schedule; the dependent ones are recovered afterwards from the equality constraints. The
layer therefore first enforces the inequality constraints on $z_i$ and then completes the
schedule with the equality constraints.

\begin{figure}[t]\centering
\includegraphics[width=\columnwidth]{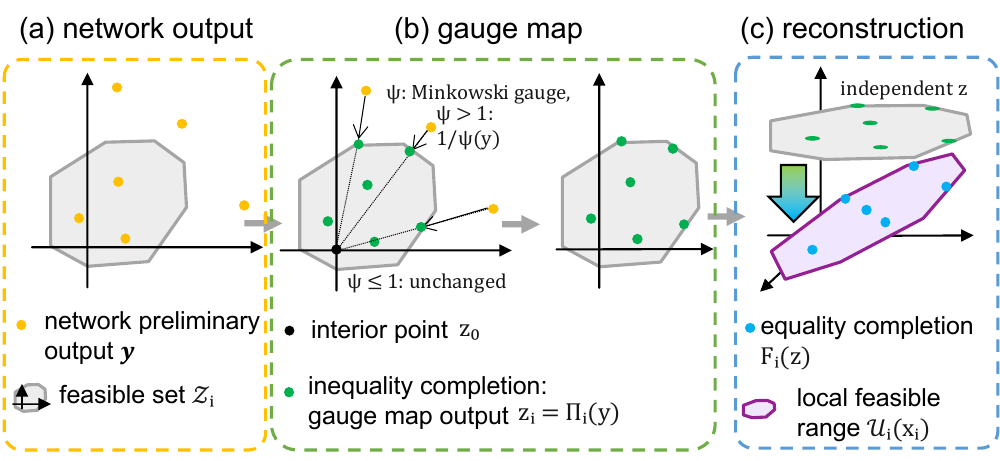}
\caption{Gauge layer: the gauge map pulls any network output into the inequality-feasible
set, and equality completion reconstructs the full schedule, so local feasibility never
depends on how well the network is trained.}
\label{fig:gauge}
\end{figure}

\textbf{Inequality constraints: gauge map} \cite{li2024gm,blanchini2008set}. Expressed in the
independent variables, the local inequality constraints form the set
\begin{equation}
\mathcal{Z}_i(x_i)=\{z:\ a_l^\top z\le b_l,\ l=1,\dots,L_i\},
\label{eq:reduced}
\end{equation}
where each of the $L_i$ rows $(a_l,b_l)$ is one linear inequality, with $b_l$ depending on
$x_i$.
Assume that the reduced set has a strict interior point $z_0$, with $b_l-a_l^\top z_0>0$ for every row. The network outputs a displacement $y$ from this point. The gauge map

\begin{equation}
\Pi_i(y)=z_0+\frac{y}{\max\{1,\psi(y)\}}
\label{eq:gauge}
\end{equation}
leaves a feasible point unchanged and pulls an infeasible point toward $z_0$ until it reaches
the boundary of $\mathcal{Z}_i$ (Fig.~\ref{fig:gauge}). For the linear constraints
\eqref{eq:reduced}, the gauge has the closed form
\begin{equation}
\psi(y)=\max_l\frac{a_l^\top y}{b_l-a_l^\top z_0},
\label{eq:psi}
\end{equation}
so the map is computed in one feed-forward pass and is differentiable almost everywhere.

\textbf{Equality constraints: variable elimination} \cite{li2023access,donti2021dc3}.
After eliminating redundant equalities, dependent variables are recovered from $z_i$ through an affine map $F_i$, giving
\begin{equation}
u_i=\big[z_i;\ F_i(z_i)\big].
\label{eq:elim}
\end{equation}
Substituting $F_i$ into the inequality constraints gives $\mathcal{Z}_i$. Thus every $z_i=\Pi_i(y)$ yields a locally feasible schedule through \eqref{eq:elim}, as illustrated in Fig.~\ref{fig:gauge}. The full optimizer is $\xi_i=[\Pi_i\circ\phi_i;\ F_i(\Pi_i\circ\phi_i)]$.

The neural optimizer is trained on reference solutions with an imitation loss plus a
normalized commitment penalty. After the last round $K$, the output is the \emph{coordinated
schedule} $\bar u_i=u_i^K$, with net power $\bar p_i=C_i\bar u_i$.

\subsection{Stage 2: Local adjustment}
Stage 2 uses each agent's flexibility and its inferred commitment gap to adjust the coordinated schedule without further communication.

\begin{definition}[Flexibility envelope and flexibility extremes]
The \emph{flexibility envelope} of agent $i$ is the set of net-power trajectories it can
realize within its own limits, $\mathcal{P}_i(x_i)=\{C_iu_i:u_i\in\mathcal{U}_i(x_i)\}$.
Its \emph{flexibility extremes} are feasible schedules constructed along directions associated with curtailment $v_i^{\mathrm C}$ and absorption $v_i^{\mathrm A}$ (consumption level), and deferral $v_i^{\mathrm D,t}$ and advance $v_i^{\mathrm V,t}$ (relative consumption in adjacent windows). This gives $2T$ directions in total: two level directions and $2(T-1)$ shift directions.

\end{definition}

Each extreme is associated with a net-power direction
$\delta\in\mathbb{R}^T$ (positive for injection): $\delta=\mathbf{1}$ for
curtailment, $-\mathbf{1}$ for absorption, $e_t-e_{t+1}$ for deferral and $e_{t+1}-e_t$ for
advance, where $\mathbf{1}$ is the all-ones vector and $e_t$ the $t$-th unit vector. By
\eqref{eq:elim}, the net power is affine in the independent variables, $p_i=G_iz_i+g_i$ with constant
$G_i$ and $g_i$, so
$d=G_i^\top\delta$ is the gradient of $\delta^\top p_i$ in the reduced coordinates. For $d\ne0$, the ray reaches the boundary at $z_0+d/\psi(d)$, which is completed by \eqref{eq:elim}; a zero direction uses $z_0$. This takes one gauge evaluation and no optimization solve. These are directional boundary schedules, not maximizers over the full flexibility envelope. Their names describe the intended adjustment: device coupling determines the actual change across windows, and only feasibility is needed for rebalancing.

\textbf{Gap inference.} After round $K$, agent $i$ still holds the messages $m_i^K$. A
linear decoder, trained offline, estimates from them the other agents' aggregate net power
$\hat p_{-i}$, giving the inferred commitment gap
\begin{equation}
\hat r_i=\bar p_i+\hat p_{-i}-s .
\end{equation}
The decoder uses the received scheduling summaries to estimate aggregate power without collecting new schedules. Because the messages describe round $K{-}1$, the estimate can differ from the actual round-$K$ commitment gap.

\textbf{Rebalancing.} Let $V_i$ stack the coordinated schedule and the $2T$ flexibility
extremes as columns, $V_i=[\bar u_i,v_i^{\mathrm{C}},v_i^{\mathrm{A}},\{v_i^{\mathrm{D},t},v_i^{\mathrm{V},t}\}_{t=1}^{T-1}]$,
and $Q_i=C_iV_i\in\mathbb{R}^{T\times(2T+1)}$. Each agent outputs the weighted sum $u_i=V_iw_i$ with
rebalancing weights
\begin{equation}
\begin{aligned}
w_{i,m}&=\max\{\sigma_{i,m}-\tau_i,\,0\},\\
\sigma_i&=w^0-Q_i^\top \Gamma_i\hat r_i-c\,\nabla_w f_i(V_iw)|_{w^0},
\end{aligned}
\label{eq:threshold}
\end{equation}
where $w^0=(1,0,\dots,0)$ places all weight on the coordinated schedule, $c\ge0$ weights the
local-cost gradient, and the threshold $\tau_i$ is the unique scalar for which
$\sum_m w_{i,m}=1$. The threshold is found in closed form by sorting the $2T{+}1$ entries of
the score $\sigma_i$, so $w_i$ lies in the probability simplex $\Delta\subset\mathbb{R}^{2T+1}$ without any
solver; entries below $\tau_i$ receive zero weight. The score combines a response to the inferred gap with a local-cost preference. The rebalancing gain $\Gamma_i=e^{\alpha_i}\Gamma$ consists of a shared matrix $\Gamma\in\mathbb{R}^{T\times T}$ and an agent-specific scale $e^{\alpha_i}$. We initialize $\Gamma$ at $2\overline{H^{-1}}$, where $H=\sum_iQ_iQ_i^\top$ and the bar averages over training instances. This initialization sets the adjustment scale; it is a heuristic, not a descent guarantee. With the neural optimizer fixed, $\Gamma$, $\alpha_i$ and $c$ are trained offline to minimize the commitment gap plus the optimality gap. Estimated gaps and simultaneous local adjustments do not guarantee a reduction for every instance; their effect is evaluated below.

Table~\ref{tab:stage2} illustrates the level and timing preferences associated with the inferred gap; device constraints and the rebalancing scores determine the actual adjustment.

\begin{table}[t]\centering
\caption{Illustrative local adjustment preferences for two windows}
\label{tab:stage2}
\begingroup
\footnotesize
\setlength{\tabcolsep}{4pt}
\renewcommand{\arraystretch}{1.0}
\begin{tabular}{@{}cll@{}}
\toprule
Example & Detected gap & Favored adjustment\\
\midrule
\begin{minipage}[c]{.19\columnwidth}\centering
\includegraphics[width=.85\linewidth]{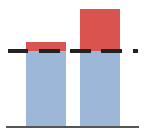}
\end{minipage}
& \begin{tabular}{@{}l@{}}Excess in both windows\\
$[\hat r_i]_1,[\hat r_i]_2>0$\\[2pt]
Larger excess later\\
$[\hat r_i]_2>[\hat r_i]_1$
\end{tabular}
& \begin{tabular}{@{}l@{}}Absorption\\(consume more)\\[2pt]
Deferral\\(consume later)\end{tabular}\\[2pt]
\midrule
\begin{minipage}[c]{.19\columnwidth}\centering
\includegraphics[width=.85\linewidth]{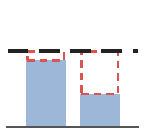}
\end{minipage}
& \begin{tabular}{@{}l@{}}Shortfall in both windows\\
$[\hat r_i]_1,[\hat r_i]_2<0$\\[2pt]
Larger shortfall later\\
$[\hat r_i]_2<[\hat r_i]_1$
\end{tabular}
& \begin{tabular}{@{}l@{}}Curtailment\\(consume less)\\[2pt]
Advance\\(consume earlier)\end{tabular}\\[2pt]
\bottomrule
\end{tabular}
\endgroup
\end{table}

\begin{proposition}[Local feasibility by construction]
For any rebalancing weights $w_i\in\Delta$, the decision $u_i=V_iw_i$ satisfies
$u_i\in\mathcal{U}_i(x_i)$, and its net power lies in the flexibility envelope
$\mathcal{P}_i(x_i)$.
\end{proposition}
\begin{proof}
Each column of $V_i$ lies in the convex set $\mathcal{U}_i(x_i)$ by \eqref{eq:elim}, and
$V_iw_i$ is their convex combination; apply $C_i$ for the net power.
\end{proof}
Per-agent extra work: $2T$ gauge-layer evaluations and a threshold; no additional communication round.

\section{Results}
\textbf{Test system.} The VPP has $N=100$ agents and $T=12$ consecutive five-minute dispatch windows, in four groups:
flexible and inflexible loads; inflexible loads with HVAC and PV; inflexible loads with EV
charging; and inflexible loads with battery storage. Load profiles come from one week of NYISO
real-time load data for New York City \cite{nyiso}; outdoor temperature and solar irradiance
for Central Park from the National Weather Service \cite{nws2023}; EV charging demand from
public charging-station utilization data \cite{borlaug2023public}; the HVAC comfort setting
from ASHRAE Standard~55 \cite{ashrae55}; and device parameters from
\cite{cui2019peer,wang2016incentivizing,li2016helos}. Instances are split into
training, validation and test sets (about 82:14:4); an independent set perturbs the local
states with 3\% Gaussian noise and is never used for tuning.

\textbf{Compared methods.} All methods run $K=5$ communication rounds. Baselines: ADMM
\cite{boyd2011admm}, with the best penalty parameter $\rho=10^{-4}$ from
$\{10^{-3},\dots,10^{-7}\}$; a penalty-based learned coordinator
with the same backbone, in which the gauge layer is replaced by loss penalties
\cite{liu2024loop}; and learned coordination alone, i.e., the same Stage~1 model with Stage~2 omitted \cite{li2023l2o}. The centralized optimum
is computed with Gurobi \cite{gurobi}.

\textbf{Metrics.} We report the \emph{commitment gap}
$E=\|\sum_i p_i - s\|_1$ (kW, averaged over test instances), also given in percent of the agents' aggregate absolute power $\sum_i\|p_i^\star\|_1$ at the optimum (about 332~kW), the maximum violation of
\eqref{eq:central_local} over all agents and instances, and the
\emph{optimality gap} $\sum(f(u)-f(u^\star))/\sum|f(u^\star)|$,
where $f(u)=\sum_if_i(u_i)$, $u^\star$ is the centralized optimum of \eqref{eq:central} computed with Gurobi \cite{gurobi}, and the
sums run over test instances. Since \eqref{eq:central_couple} holds only up to $E$, the
optimality gap compares costs and is not a certified suboptimality bound.

\begin{table}[t]\centering
\caption{Main results ($E$: commitment gap)}
\label{tab:main}
\setlength{\tabcolsep}{2.5pt}
\begin{tabular}{lcccc}
\toprule
Method & $E$, kW (\%) & Max viol. & Loc.\ feas. & Opt.\ gap\\
\midrule
ADMM~\cite{boyd2011admm} & 262.7 (79) & 0 & yes & $-8.45\%^\dagger$\\
Penalty~\cite{liu2024loop} & 12.02 (3.6) & 7.37 & \textbf{no} & $-16.1\%^\ddagger$\\
Coord.\ only~\cite{li2023l2o} & 16.88 (5.1) & 0 & yes & 3.93\%\\
\textbf{LOOP-MAC} & \textbf{2.48 (0.75)} & 0 & \textbf{yes} & \textbf{3.07\%}\\
\midrule
LOOP-MAC, indep.\ set & 3.77 (1.13) & 0 & yes & 5.23\%\\
\bottomrule
\end{tabular}
\\[2pt]{\footnotesize ADMM: $\rho=10^{-4}$; Penalty: soft constraints; Coord.\ only: learned coordination alone; Loc.\ feas.: local limits \eqref{eq:central_local} met. $^\dagger$Below the optimum only because the commitment gap exceeds 75\%. $^\ddagger$Below the optimum because local limits are violated.}
\end{table}

\textbf{Main comparison.} In Table~\ref{tab:main}, ADMM
is locally feasible but far from the commitment. The penalty method reaches a
small gap only by violating local limits. Optimality gaps must be read together with $E$:
ADMM appears cheaper than the optimum only because it misses the commitment.
We also tested ML-ADMM \cite{biagioni2022admm,mak2023admm}, which initializes ADMM with learned
predictions. Its initial schedules are close to each agent's optimum, but the multiplier
updates that enforce the commitment do not adapt well to this initialization, so its
five-round commitment gap remained far larger than that of ADMM.
LOOP-MAC attains the smallest commitment gap among locally feasible methods: the local
adjustment reduces the gap of learned coordination alone substantially while also lowering
the optimality gap. On the independent set (last row), LOOP-MAC remains locally feasible,
with $E$ rising from 0.75\% to 1.13\%. Because the inferred gap is an estimate, the local
adjustment increases $E$ on about 10\% of independent instances, by at most 1.5~kW (0.45\%).

\begin{table}[t]\centering
\caption{Commitment gap $E$, kW (\%), versus communication rounds, original data}
\label{tab:rounds}
\setlength{\tabcolsep}{4pt}
\begin{tabular}{crrr}
\toprule
Rounds $K$ & ADMM & Coord.\ only & LOOP-MAC\\
\midrule
5 & 262.7 (79) & 16.88 (5.1) & 2.48 (0.75)\\
10 & 146.8 (44) & 16.99 (5.1) & 2.21 (0.66)\\
\bottomrule
\end{tabular}\\[2pt]
{\footnotesize ADMM: $\rho=10^{-4}$; local adjustment trained separately for each $K$.}
\end{table}

\textbf{Number of rounds.} Table~\ref{tab:rounds} shows little change in learned coordination alone and a modest improvement in LOOP-MAC as the communication budget increases. ADMM improves more but retains a larger commitment gap. The comparison supports the five-round setting for LOOP-MAC in this case study.

\textbf{Ablation.} Table~\ref{tab:abl} separates the local adjustment from its learned calibration. The initial adjustment improves commitment tracking over learned coordination alone, and training the rebalancing parameters further reduces the gap. The ordering is consistent across the original and independent sets.

\begin{table}[t]\centering
\caption{Ablation of the local adjustment, $E$, kW (\%)}
\label{tab:abl}
\setlength{\tabcolsep}{3pt}
\begin{tabular}{lcc}
\toprule
Variant & Original & Indep.\\
\midrule
Coord.\ only & 16.88 (5.1) & 18.08 (5.4)\\
+ adjustment, initial gain & 6.95 (2.1) & 7.60 (2.3)\\
+ adjustment, trained gain (\textbf{LOOP-MAC}) & \textbf{2.48 (0.75)} & \textbf{3.77 (1.13)}\\
\bottomrule
\end{tabular}
\end{table}

\begin{table}[t]\centering
\caption{Per-agent online computation time}
\label{tab:time}
\setlength{\tabcolsep}{4pt}
\begin{tabular}{lrr}
\toprule
Component & LOOP-MAC & ADMM\\
\midrule
One coordination round & 158~$\mu$s & 0.33~s\\
Local adjustment & 293~$\mu$s & ---\\
Total, five rounds and adjustment & 1.1~ms & 1.65~s\\
\bottomrule
\end{tabular}\\[2pt]
{\footnotesize Single-thread CPU computation; total includes local adjustment for LOOP-MAC.}
\end{table}

\textbf{Computation time.} Table~\ref{tab:time} shows that learned coordination, including the local adjustment, requires substantially less computation than the recorded ADMM local solves.

\section{Conclusion}
LOOP-MAC combines gauge-based learned coordination with a local adjustment that uses messages already received. The adjustment preserves local feasibility through convex combinations and, on the 100-agent VPP, reduces the five-round commitment gap by 85\% relative to learned coordination alone while lowering the optimality gap. Local feasibility is guaranteed under the convex device model; commitment tracking remains approximate and depends on the quality of the inferred gap.

\section*{Acknowledgment}
The authors used AI-based tools (Claude, Anthropic; Codex, OpenAI) to assist with language
editing and code development. All content was reviewed and verified by the authors, who take
full responsibility for it.

\bibliographystyle{IEEEtran}

\begin{thebibliography}{99}
\bibitem{chen2023tracking} Y. Chen, J. Chen, C. Ge, W. Zhong, and M. Liu, ``Scheduled power tracking control of the virtual power plant for its internal contingency considering the communication delay and the unit capacity limitation,'' \emph{Electr. Power Syst. Res.}, vol. 221, p. 109402, 2023.
\bibitem{chen2024disagg} Q. Chen, R. Lyu, H. Guo, and X. Su, ``Real-time operation strategy of virtual power plants with optimal power disaggregation among heterogeneous resources,'' \emph{Appl. Energy}, vol. 361, p. 122876, 2024.
\bibitem{wang2026l2o} Q. Wang and Z. Li, ``Learning-to-optimize infused decentralized disaggregation for multi-entity technical VPP considering equity and privacy,'' \emph{Appl. Energy}, vol. 405, p. 127226, 2026.
\bibitem{molzahn2017survey} D. K. Molzahn, F. D\"orfler, H. Sandberg, S. H. Low, S. Chakrabarti, R. Baldick, and J. Lavaei, ``A survey of distributed optimization and control algorithms for electric power systems,'' \emph{IEEE Trans. Smart Grid}, vol. 8, no. 6, pp. 2941--2962, 2017.
\bibitem{boyd2011admm} S. Boyd, N. Parikh, E. Chu, B. Peleato, and J. Eckstein, ``Distributed optimization and statistical learning via the alternating direction method of multipliers,'' \emph{Found. Trends Mach. Learn.}, vol. 3, no. 1, pp. 1--122, 2011.
\bibitem{zeng2022rl} S. Zeng, A. Kody, Y. Kim, K. Kim, and D. K. Molzahn, ``A reinforcement learning approach to parameter selection for distributed optimal power flow,'' \emph{Electr. Power Syst. Res.}, vol. 212, p. 108546, 2022.
\bibitem{feng2023admm} C. Feng, K. Zheng, Y. Zhou, P. Palensky, and Q. Chen, ``Update scheduling for ADMM-based energy sharing in virtual power plants considering massive prosumer access,'' \emph{IEEE Trans. Smart Grid}, vol. 14, no. 5, pp. 3961--3975, 2023.
\bibitem{biagioni2022admm} D. Biagioni, P. Graf, X. Zhang, A. S. Zamzam, K. Baker, and J. King, ``Learning-accelerated ADMM for distributed DC optimal power flow,'' \emph{IEEE Control Syst. Lett.}, vol. 6, pp. 1--6, 2022.
\bibitem{mak2023admm} T. W. K. Mak, M. Chatzos, M. Tanneau, and P. Van Hentenryck, ``Learning regionally decentralized AC optimal power flows with ADMM,'' \emph{IEEE Trans. Smart Grid}, vol. 14, no. 6, pp. 4863--4876, 2023.
\bibitem{li2023l2o} M. Li, S. Kolouri, and J. Mohammadi, ``Learning to optimize distributed optimization: ADMM-based DC-OPF case study,'' in \emph{Proc. IEEE PES General Meeting}, 2023, pp. 1--5.
\bibitem{noah2025unfold} Y. Noah and N. Shlezinger, ``Distributed learn-to-optimize: Limited communications optimization over networks via deep unfolded distributed ADMM,'' \emph{IEEE Trans. Mobile Comput.}, vol. 24, no. 4, pp. 3012--3024, 2025.
\bibitem{saravanos2025deep} A. D. Saravanos, H. Kuperman, A. Oshin, A. T. Abdul, V. Pacelli, and E. A. Theodorou, ``Deep distributed optimization for large-scale quadratic programming,'' in \emph{Proc. ICLR}, 2025.
\bibitem{liu2024loop} X. Liu, Y. Lu, A. Abbasi, M. Li, J. Mohammadi, and S. Kolouri, ``Teaching networks to solve optimization problems,'' \emph{IEEE Access}, vol. 12, pp. 17102--17113, 2024.
\bibitem{parshakova2025mra} T. Parshakova, Y. Bai, G. van Ryzin, and S. Boyd, ``Multiple-response agents: Fast, feasible, approximate primal recovery for dual optimization methods,'' arXiv:2503.12221, 2025.
\bibitem{donti2021dc3} P. L. Donti, D. Rolnick, and J. Z. Kolter, ``DC3: A learning method for optimization with hard constraints,'' in \emph{Proc. ICLR}, 2021.
\bibitem{liang2023homeo} E. Liang, M. Chen, and S. Low, ``Low complexity homeomorphic projection to ensure neural-network solution feasibility for optimization over (non-)convex set,'' in \emph{Proc. ICML}, PMLR 202, 2023, pp. 20623--20649.
\bibitem{li2023access} M. Li, S. Kolouri, and J. Mohammadi, ``Learning to solve optimization problems with hard linear constraints,'' \emph{IEEE Access}, vol. 11, pp. 59995--60004, 2023.
\bibitem{li2024gm} M. Li and J. Mohammadi, ``Toward rapid, optimal, and feasible power dispatch through generalized neural mapping,'' in \emph{Proc. IEEE PES General Meeting}, 2024, pp. 1--5.
\bibitem{blanchini2008set} F. Blanchini and S. Miani, \emph{Set-Theoretic Methods in Control}. Boston, MA, USA: Birkh\"auser, 2008.
\bibitem{gurobi} Gurobi Optimization, LLC, \emph{Gurobi Optimizer Reference Manual}, 2026.
\bibitem{nyiso} New York Independent System Operator, ``Real-time load data,'' 2023. [Online]. Available: https://www.nyiso.com/real-time-dashboard
\bibitem{ashrae55} ASHRAE, \emph{Thermal Environmental Conditions for Human Occupancy}, ANSI/ASHRAE Standard 55, 1992.
\bibitem{nws2023} National Weather Service, ``Weather data for New York City's Central Park,'' 2023. [Online]. Available: https://w1.weather.gov/data/obhistory/KNYC.html
\bibitem{borlaug2023public} B. Borlaug, F. Yang, E. Pritchard, E. Wood, and J. Gonder, ``Public electric vehicle charging station utilization in the United States,'' \emph{Transp. Res. Part D}, vol. 114, p. 103564, 2023.
\bibitem{cui2019peer} S. Cui, Y.-W. Wang, and J.-W. Xiao, ``Peer-to-peer energy sharing among smart energy buildings by distributed transaction,'' \emph{IEEE Trans. Smart Grid}, vol. 10, no. 6, pp. 6491--6501, 2019.
\bibitem{wang2016incentivizing} H. Wang and J. Huang, ``Incentivizing energy trading for interconnected microgrids,'' \emph{IEEE Trans. Smart Grid}, vol. 9, no. 4, pp. 2647--2657, 2018.
\bibitem{li2016helos} G. Li, D. Wu, J. Hu, Y. Li, M. S. Hossain, and A. Ghoneim, ``HELOS: Heterogeneous load scheduling for electric vehicle-integrated microgrids,'' \emph{IEEE Trans. Veh. Technol.}, vol. 66, no. 7, pp. 5785--5796, 2017.
\end{thebibliography}
\linespread{0.92}\def\IEEEbibitemsep{0pt}

\end{document}